\documentclass[11pt]{article}

\usepackage[preprint]{acl}
\usepackage{enumitem}

\usepackage{xurl}

\usepackage{times}
\usepackage{latexsym}
\usepackage{todonotes}

\usepackage[T1]{fontenc}

\usepackage[utf8]{inputenc}

\usepackage{microtype}

\usepackage{inconsolata}

\usepackage{graphicx}
\usepackage{expex, amsmath, amssymb, amsfonts, subcaption, dblfloatfix}
\usepackage{lipsum}

\title{Syntactic Belief Update\\as the Driver of Garden Path Processing Difficulty}

\author{
    Alan Zhou\thanks{Equal contribution.} \\ Johns Hopkins University \\ \texttt{azhou23@jhu.edu}
    \And   
    Milo\v{s} Stanojevi\'{c}\footnotemark[1] \\ University College London \\ \texttt{m.stanojevic@ucl.ac.uk} 
    \And
    John T. Hale \\ Johns Hopkins University \\ \texttt{jthale@jhu.edu}
}

\renewenvironment{quote}{%
  \list{}{\rightmargin\leftmargin \topsep=0pt}
  \item\relax
}{%
  \endlist
}

\begin{document}
\maketitle
\begin{abstract}
Garden path sentences present a processing difficulty for humans --- the sentence prefix leads the listener towards one interpretation, until the listener hears a critical word that shows that the initial interpretation was wrong.
Lexical surprisal, a measure that usually predicts sentence processing difficulty quite well, fails to provide good predictions for garden path sentences.

We propose an alternative that actively predicts a distribution over syntactic trees (its syntactic belief) and updates that distribution after each new word. If a processor is led down a garden path, syntactic beliefs will be wrong, requiring a large update at the critical word. The magnitude of the update is measured with a R\'enyi divergence. Crucially, this metric is dependent on lexical items, but is distinct from lexical probability and measures a degree of belief update over non-lexical syntactic structure only.
This Syntactic Belief Update provides a better fit to human reading time data on garden path sentences,  suggesting a new research direction examining non-lexical alternatives to surprisal for psycholinguistics.
\end{abstract}



\section{Introduction}
A prevailing view of human sentence processing casts comprehension in an information-theoretic treatment of lexical items, relating the difficulty of processing an oncoming word to its \textit{surprisal}, i.e. its negative logarithm probability of appearing given the prior context \cite{haleProbabilisticEarleyParser2001, levyExpectationbasedSyntacticComprehension2008}. In its strongest form \citep{levyExpectationbasedSyntacticComprehension2008, smithEffectWordPredictability2013}, that view proposes surprisal to be a potential \textit{causal bottleneck} through which all processing is mediated, which can explain all or most of sentence comprehension difficulty in humans. Indeed, surprisal estimates extracted out of language models have enjoyed a wide range of empirical support, providing remarkably strong predictions of human reading time \citep{smithEffectWordPredictability2013, wilcoxLanguageModelQuality2023, wilcoxTestingPredictionsSurprisal2023} and human neural activity \citep{bhattasaliUsingSurprisalFMRI2021, brodbeckParallelProcessingSpeech2022, gillisNeuralMarkersSpeech2021, stanojevicModelingStructureBuildingBrain2023}. These impressive results have fueled a growing wave of work viewing word predictability as the main driver of human sentence processing difficulty \cite{smithEffectWordPredictability2013,schrimpfNeuralArchitectureLanguage2021, wilcoxLanguageModelQuality2023, wilcoxTestingPredictionsSurprisal2023}

However, a line of recent work \cite{vanschijndelModelingGardenPath2018, arehalliSyntacticSurprisalNeural2022,huangLargescaleBenchmarkYields2024} suggests that the remarkable success of surprisal on explaining human processing difficulty on broad-coverage data does not extend to processing difficulty on specific syntactic phenomena known as garden paths. These are sentences, such as \textit{The old man the boat}, in which the most likely incremental interpretation of a sentence is incompatible with the full parse (e.g. \textit{man} is a verb, not a noun), resulting in costly reanalysis or comprehension failure. Surprisal estimates from language models drastically underpredict the slowdown that humans experience when reading such constructions, and additionally fail to capture the relative grades of difficulty that humans experience over different types of garden paths \cite{prasadHowMuchHarder2019, huangLargescaleBenchmarkYields2024}. Such results suggest that garden path difficulty is indicative of some additional mechanism of the human sentence processing algorithm that is not fully captured by lexical predictability.



One natural extension is to enrich surprisal computation with syntactic structure. This approach was known since the introduction of surprisal to psycholinguistics \citep{haleProbabilisticEarleyParser2001}. However, many of these approaches reduce to lexical surprisal \citep{levyExpectationbasedSyntacticComprehension2008} and mispredict the garden path effect \citep{vanschijndelModelingGardenPath2018, arehalliSyntacticSurprisalNeural2022, huangLargescaleBenchmarkYields2024}.


Instead, we investigate the hypothesis that garden path processing is best explained by the specific psycholinguistic notion of \textit{syntactic reanalysis} \cite{frazierMakingCorrectingErrors1982, fodorReanalysisSentenceProcessing1998, lewisReanalysisLimitedRepair1998}, a costly procedure in which the human parser must abandon (or downgrade the score of) a syntactic interpretation it has already committed to when confronted with conflicting evidence. Taking inspiration from Bayesian models of sentence processing \cite{narayananBayesianModelsHuman2022, narayananBayesianModelPredicts2002}, we formalize syntactic reanalysis as  the degree of the update to syntactic belief in information theoretic terms, i.e. the predicted probability distribution over delexicalized parse trees, after new evidence (the next word) is observed. Crucially, our proposed metric measures belief update over \textit{unlexicalized structures}, and is distinct from lexical predictability--- the difficulty of syntactic belief update is influenced by the upcoming word, but a word being more predictable in the context does not directly determine whether the degree of the syntactic belief update will be larger or smaller.
While lexical probabilities do have an effect in general sentence processing, we argue that garden path difficulty is best explained by the changes in delexicalized syntactic beliefs.
Our code  is available on GitHub.\footnote{\url{https://github.com/atzhou8/discriminative_incremental_parsing}}


\section{Background}
\label{sec:background}
\subsection{Garden Paths}
Humans exhibit significant slowdowns when reading sentences with temporary syntactic ambiguities, known as garden paths \cite{bever1970}. These are grammatical sentences in which the most likely initial interpretation of a sentence is incompatible with the final parse, leading comprehenders ``down the garden path''. The remarkable difficulty we face with these otherwise syntactically well-formed sentences gives us a window into the specific mechanisms of the human sentence processing algorithm. As such, the question of how garden paths are processed by humans has been a long-standing problem in psycholinguistics since the 1970s (\citet{bever1970, frazierMakingCorrectingErrors1982,  fodorReanalysisSentenceProcessing1998, pritchettGardenPathPhenomena1987}, among others).

Two facts about garden path processing are particularly relevant. First, the slowdowns that people exhibit are most pronounced at the word at which the structure becomes disambiguated--- as well as the words immediately following it--- in an area often called the \textit{critical region}. Second, the difficulty of processing garden path sentences is attributable to ambiguities in their syntactic structure, and some ambiguities are easier to process than others. Here we show three garden path types and their unproblematic equivalents for which the largest amount psycholinguistic data exists: Noun Phrase vs. Sentential Complement (NP/S; \citealp{frazier1979comprehending}), Noun Phrase vs. Zero Complement (NP/Z; \citealp{frazier1979comprehending}), and Main Verb vs. Reduced Relative (MV/RR; \citealp{bever1970}). Each of these examples are sourced from \citet{huangLargescaleBenchmarkYields2024}, with their  critical regions rendered in italics.
\lingset{belowexskip=1pt,aboveexskip=1pt}
\pex Noun Phrase/Sentential Complement \textbf{(NP/S)}\label{ex:nps}
\a The girl found the lamb \textit{remained relatively calm.}\trailingcitation{(\textit{garden path})}
\a The girl found \textbf{that} the lamb \textit{remained relatively calm.}\trailingcitation{(\textit{control})}
\xe
\pex Noun Phrase/Zero Complement \textbf{(NP/Z)}\label{ex:npz}
\a When the girl attacked the lamb \textit{remained relatively calm.}\trailingcitation{(\textit{garden path})}
\a When the girl attacked \textbf{(,)} the lamb \textit{remained relatively calm.}\trailingcitation{(\textit{control})}
\xe
\lingset{belowexskip=1pt,aboveexskip=1pt}
\pex Main Verb/Reduced Relative \textbf{(MV/RR)}\label{ex:mvrr}
\a The girl fed the lamb \textit{remained relatively calm.}\trailingcitation{(\textit{garden path})}
\a The girl \textbf{who was} fed the lamb \textit{remained relatively calm.}\trailingcitation{(\textit{control})}\\
\xe
The profile of slowdown forms a consistent hierarchy of difficulty: MV/RR is associated with the highest slowdown, followed by NP/Z, followed by NP/S \cite{huangLargescaleBenchmarkYields2024, prasadHowMuchHarder2019, sturtStructuralChangeReanalysis1999, grodnerRepairBasedReanalysisSentence2003a}.

One explanation for the timing and relative ranking of garden path difficulty comes from a family of psycholinguistic theories centered around \textit{syntactic reanalysis}, in which garden path difficulty is explained by the necessity of a costly procedure in which comprehenders revise their syntactic expectations about a sentence upon reaching the critical region. Reanalysis plays a central role for many psycholinguistic theories \citep{pritchettGardenPathPhenomena1987, ferreiraRecoveryMisanalysesGardenpath1991, lewisReanalysisLimitedRepair1998, grodnerRepairBasedReanalysisSentence2003a}, and multi-stage models that assume reanalysis have been show to better explain the slowdowns observed during garden paths \citep{paapeEstimatingTrueCost2022, paapeDeconstructingSentenceDisambiguation2026}.




\subsection{Surprisal}
The surprisal of a word $w_i$ given its preceding context $w_1\dots w_{i-1}$ is defined  as the negative logarithm of its conditional probability:\\
    $surp(w_i\!\!\mid\!\!w_{1}\dots w_{i-1})=-\log p(w_i\mid w_1\!\!\dots \!\!w_{i-1})$.

Surprisal plays an important role in computational psycholinguistics as a complexity metric for human sentence processing difficulty \cite{haleProbabilisticEarleyParser2001, levyExpectationbasedSyntacticComprehension2008}.
Surprisal is highly predictive of human psychometrics like reading time and neural activity \cite{smithEffectWordPredictability2013, gillisNeuralMarkersSpeech2021, wilcoxLanguageModelQuality2023, wilcoxTestingPredictionsSurprisal2023,stanojevicModelingStructureBuildingBrain2023}. 

%
%
\citet{levyExpectationbasedSyntacticComprehension2008} proved that lexical surprisal can be reduced to KL divergence between distributions covering upcoming words $X_{i}\dots X_{n}$ and an optional latent variable $Y$, conditioned on the prefix $w_1 \dots w_{i-1}$ before and after observing a word $w_i$:
\begin{align}\label{eq:surp-kl}
    &surp(w_i\mid w_1\dots w_{i-1}) \nonumber \\
      &= -\log\ p(w_i\mid w_1\dots w_{i-1}) \nonumber \\
    &= D_{KL}( \nonumber \\
    &\phantom{= D} p\left(X_i, \dots, X_n, Y \mid w_1\dots w_{i-1}, X_i\!\!=\!\!w_i \right)\mid\mid \nonumber \\
    &\phantom{= D} p\left(X_i, \dots, X_n, Y \mid w_1\dots w_{i-1} \right) \nonumber\\
    &\phantom{==})
\end{align}

Surprisal's interpretation as a psychometric predictor, then, is the effort associated with \textit{disconfirming} a set of previously-entertained lexical predictions in light of the information provided by a new  word. In this formulation, the latent variable $Y$, which can be syntactic structure, is optional and influences the next word's probability. However, in the end, surprisal is still representing only the negative logarithm of the next word's probability under the given model. Surprisal is acting as a \textit{causal bottleneck} that abstracts away from the syntactic latent variable (if present) to compute a lexical measure of processing difficulty.

One consequence of this causal bottleneck is surprisal's ``representation agnosticism''-- regardless of the model computing next-word probabilities, surprisal implicitly factors in the belief update over all types of latent structures that a comprehender must consider. This has been portrayed as a desirable property in some work \citep{futrellLossyContextSurprisalInformationTheoretic2020}, but has been criticized by others as obfuscating the cognitive role of these latent representations beyond their influence on downstream lexical probabilities \citep{slaatsWhatsSurprisingSurprisal2025, buxo-lugoSurprisalNotTheory2026}.

This tension is especially relevant for garden paths, where the update of \textit{syntactic beliefs} specifically is generally taken to be the main source of difficulty. Surprisal underpredicts the magnitude of garden path difficulty, and fails to capture the \mbox{NP/S < NP/Z < MV/RR} asymmetry \citep{vanschijndelSinglestagePredictionModels2021, arehalliSyntacticSurprisalNeural2022, huangLargescaleBenchmarkYields2024}, despite evidence that the latent representations of neural language models encode reanalysis at least some of the time \citep{Li24_incremental}. We hypothesize this failure to be attributed to this exact representation agnosticism, which obfuscates the specific difficulty of updating syntactic beliefs.



\section{Our proposal: Syntactic Belief Update}
\label{sec:our:proposal}

We propose an alternative complexity metric in which the unlexicalized syntactic structure is the central determinant of processing effort, rather than just one of many latent factors mediated through a causal bottleneck to determine lexical probabilities.

\subsection{Formal Definition}
\label{sec:our:proposal:syntactic:belief:update}

We define \textit{Syntactic Belief} for a prefix string $w_1\dots w_{i-1}$ to be a conditional probability distribution over syntactic trees: $q(Y \mid w_1\dots w_{i-1})$. After a new word is observed $w_i$ it is treated as new evidence that triggers a Bayesian update of the Syntactic Belief to get a new belief $p(Y \mid w_1\dots w_i)$. The magnitude of this \textit{Syntactic Belief Update (SBU)} can be measured with a divergence:
\begin{align}
    S&_{BU}(w_1\dots w_{i-1}, w_i) \label{eq:sbu:kl} \\ 
     & = D\left(p(Y \mid w_1\dots w_{i}) \mid\mid q(Y \mid w_1\dots w_{i-1})\right) \nonumber
\end{align}

Formally, any divergence, not just KL divergence, could be a valid candidate as long as it satisfies core properties of a divergence of probability distributions \citep{information:geometry:amari2000methods}:
\begin{description}
    \item[\textbf{non-negativity}] $D(p \mid\mid q) \geq 0$ and
    \item[\textbf{identity of indiscernibles}] $D(p\!\!\mid\mid\!\!q)\!=\!0\!\!\iff\!\!p\!\!=\!\!q$.
\end{description}
Intuitively these properties are aligned with what we expect from cognitive processing effort: the effort cannot be negative and if there is no change to the distribution we expect zero processing cost.

Different divergences would have a different prediction of the degree and a different explanation of the processing effort. Here we consider a family of divergences called R\'enyi divergences which are parametrized by a single parameter $\alpha$ satisfying \mbox{$0 < \alpha < \infty$ and $\alpha \neq 1$} \citep{renyi-entropy-original}:
\begin{align}
    D_{\alpha}(p\!\mid\mid\!q) = \frac{1}{\alpha-1}\log \sum_{x} p(x)\!\left(\!\frac{p(x)}{q(x)}\!\right)^{\alpha-1}
\end{align}

Using a family of divergences allows us to explore a class of divergences in a structured way. KL divergence is a special case of R\'enyi divergence when $\alpha$ approaches $1$. For $\alpha>1$ the divergence puts more emphasis on the comparison of heads of the distributions, while $0<\alpha<1$ puts more emphasis on tails.

We note that while some divergences are symmetric,%
\footnote{e.g. Bhattacharyya distance when $\alpha=0.5$}
this is generally not the case.
Different directionality changes both the result and the interpretation of the proposal. 
We can interpret $D_{\alpha}(p \mid\mid q)$ as a measure of how many bits are wasted on average by using a codebook estimated on $q$ to encode samples from $p$, with the price of code length growing exponentially with $\alpha$ \citep{CAMPBELL1965423}. 
The right directionality in our context is therefore to set $q$ as the distribution before seeing the next word and $p$ as the distribution after. 

\subsection{Relationship to Surprisal}
The primary distinction between the divergence in surprisal (Eq. \ref{eq:surp-kl}) and the divergence in SBU (Eq. \ref{eq:sbu:kl}) is that the former is computed over the joint distribution of words $X$ and non-lexical structure $Y$, while the latter is computed only over the non-lexical structure $Y$. 

\citet{viglyComprehensionEffortCost2025} show that this belief update can be taken as part of a decomposition of surprisal:
\begin{align*}
su&rp(w_i) = \nonumber\\
&D_{KL}\left(p(Y \mid w_1 \dots w_{i}) \mid\mid p(Y \mid w_1\dots w_{i-1})\right) \nonumber\\
&+ \underset{P(Y \mid  w_1 \dots w_i)}{\mathbb{E}}\left[ \log \frac{1}{p(w_{i} \mid  Y, w_{1}, \dots w_{i-1})}\right] 
\end{align*}
where the first term is a KL divergence that is analogous to the belief update that we propose, and the second term, called reconstruction information, corresponds to the expected surprisal of word $w_i$ under the belief distribution post-update.

This implies two facts about Syntactic Belief Update for the special case when it is calculated as a KL divergence (i.e. $\alpha\rightarrow 1$). First, it is upper-bounded by surprisal, as both belief update and reconstruction information are non-negative. Second, the relationship between SBU and lexical surprisal is not necessarily monotonic---a word being more predictable than an alternative does not determine if its belief update will be higher or lower relative to the belief update of the other word. Instead, the main contribution of word predictability is in the reconstruction information term, which directly considers the predictability of $w_i$ after belief update has taken place. Intuitively, two words with similar syntactic belief updates can have very different surprisals due to reconstruction information (e.g. \textit{The dog chased the \textbf{cat}} vs. \textit{The dog chased the \textbf{table}}).

\subsection{Incremental Dependency Parsing Instantiation}

We instantiate the formal framework of SBU in an incremental parser over \textit{non-projective} dependency trees, which capture important syntactic relations such as bi-lexical relations, long-range dependencies, crossing dependencies, and for which a large number of treebanks are available through the Universal Dependencies project (UD, \citealp{nivreUniversalDependencies2017}). 
While some psycholinguistic work on processing difficulty has used dependency structures, to the best of our knowledge, prior work has been limited to projective dependency trees \citep{noji-miyao-2014-left,berta:franzluebbers-etal-2024-multipath,left:corner:Dunagan2026.04.20.719609}. Our work is the first to model human sentence processing with fully incremental non-projective dependency parsing.

In order to incorporate any syntactic representation $Y$ in our Syntactic Belief Update framework we need a way of computing $q(Y \mid w_1\dots w_{i-1})$, $p(Y \mid w_1\dots w_{i})$, and $D_{\alpha}(p \mid\mid q)$ efficiently. 
In order to compute $D_\alpha$, we must have $p$ and $q$ assign probability over the exact same set of syntactic structures $Y$, despite being conditioned on different prefixes. 
To ensure this, we opt for a very simple solution that fits the experimental setup to which human subjects were exposed when garden path data was collected:

\begin{quote}
``The materials were presented in the self-paced reading paradigm.
In this paradigm, the words of the sentence are first obscured by dashes.
The participant presses a key to reveal the words of the sentence one at a time, with each word replaced by dashes once the participant moves on from it.
''
\citep{huangLargescaleBenchmarkYields2024}
\end{quote}

In this self-paced reading setup \cite{Just1982-cq}, human participants knew the length of the sentence from the start by seeing the number of dashes. This allowed them to start forming hypotheses of the tree structure that covers all $n$ words. 
Thus, in our setup we will treat $Y$ as a set of all parse trees of length $n$ even in the cases when we have seen only a prefix of $i<n$. 

However, if we have a prefix of length $i<n$, to get the full probability of parse tree of size $n$ we would need to marginalize over intractably many suffixes ($v^{n-i}$ for vocabulary size  $v$).
Instead of marginalization during test time we opt for directly learning a marginalized model at training time -- at each training instance we randomly sample position $i$ in a sentence and replace all words $w_i \dots w_n$ with \textit{MASK} tokens. 
This way of modeling $p(Y\mid w_1\dots w_{i-1})$ is very close to the way human subjects in \citet{huangLargescaleBenchmarkYields2024} were exposed to the stimuli
where dash served as a \textit{MASK}.

To assign probability over non-projective dependency trees, we build a Conditional Random Field with an edge-factored model \citep{kooStructuredPredictionModels2007}. 
\begin{align}\label{eq:tree_prob}
    p(t)
    = \frac{\phi(t)}{\sum_{t'\in T} \phi(t')}
    = \frac{\prod_{e\in t}\phi(e)}{\sum_{t' \in T} \prod_{e'\in t'} \phi(e')}
\end{align}
Here $\phi(t)$ and $\phi(e)$ are non-negative potentials for tree $t$ and edge $e$ respectively. For any given tree $t$, computing the numerator is trivial. Most of the computational complexity is in the denominator (partition function) that needs to sum over all possible non-projective dependency trees. To compute the partition function, we use \citeposs{kooStructuredPredictionModels2007} extension of the Matrix-Tree Theorem (MTT; \citealt{tutte1984graphtheory}). With MTT we can compute the partition function as a determinant of grounded Laplacian matrix of all pairwise potentials.

We derive a new method to compute R\'enyi Divergence for any $\alpha$ in any graphical model by relying only on the efficient computation of the partition functions for $p$, $q$ and ${p^\alpha q^{1-\alpha}}$ (full derivation is provided in Appendix \ref{sec:app:renyi}):  

\begin{align}
    D_{\alpha}(p \mid\mid q) = \log Z_{q} + \frac{\log Z_{p^{\alpha}q^{1-\alpha}} - \alpha \log Z_{p}}{\alpha-1} \nonumber
\end{align}

To obtain the potentials $\phi(i, j)$, we build a scoring model by fine-tuning RoBERTa (\texttt{roberta-large} from HuggingFace; \citealp{roberta}, \citealp{huggingface}) with added prediction through a bi-affine attention layer \citep{dozatDeepBiaffineAttention2017} to maximize Equation~\ref{eq:tree_prob}. Our model is incremental but also non-autoregressive, i.e. it recomputes potentials for each new prefix. This is needed to avoid Label Bias present in all discriminative incremental models that make score of past decisions independent of future words \citep{stanojevicMaxMarginIncrementalCCG2020}.
Training details are in the Appendix~\ref{sec:app:training}.

\subsection{A Garden Path Example}
We apply our incremental dependency parser to modeling human reading behavior over garden path sentences. 
Consider a sentence such as \textit{``The old man the boat.''}
To obtain the predicted difficulty at the disambiguating word \textit{the}, we obtain full parse distributions for consecutive masked prefixes from our parser:
\begin{enumerate}
    \small \item  \texttt{The old man <mask> <mask> <mask>}
    \small \item  \texttt{The old man the <mask> <mask>}
\end{enumerate}
Here, each \texttt{<mask>} corresponds to a masked future token, including the period. We expect the distribution for the first prefix to be strongly centered around the preferred reading, in which \textit{man} is a noun. At the second prefix this distribution should shift towards the dispreferred interpretation in which \textit{man} is the matrix verb. 
We take the degree of this shift to be indicative of garden path difficulty, which we measure using a R\'enyi Divergence between the two prefix distributions. Higher $\alpha$ further penalizes shifts along the heads of distributions, predicting even higher processing difficulty in cases where the two distributions are strongly focused at different parses.


\section{Relation to Prior Syntactic Surprisal Proposals}
\label{sec:other:proposals}

 Syntactic Belief Update, as formulated here, is composed of (1) a specific abstract representation that is updated and (2) a specific way of measuring the size of update. Here we compare our proposal to the other works that are most related to ours either by the representation or by the measure of update.

\paragraph{Generative Syntactic Surprisal}

There have been many proposals in the past that model a joint probability of trees and words and measure processing difficulty in terms of the changes to the joint distribution after the next word is observed \citep{haleProbabilisticEarleyParser2001, vanschijndelModelingGardenPath2018, dyerRecurrentNeuralNetwork2016,hale-etal-2018-finding}. 
Following \citet{levyExpectationbasedSyntacticComprehension2008}, all these joint probability models reduce to a lexical surprisal when we compute KL divergence after observing the next word.
Empirically, this type of syntactically-informed lexical surprisal makes similar predictions about processing difficulty as models without an explicit latent variable \cite{vanschijndelModelingGardenPath2018}.


\paragraph{Deterministic syntax} Some theorists score probability distributions over syntax-like structures, but assume a deterministic mapping from words to structures. \citet{demberg2008data} test an unlexicalized PCFG distribution over syntactic trees that end with POS tags (not words). For them, the mapping of words to POS tags was deterministic (and incremental) which is suboptimal, especially with garden path sentences (e.g. ``the old \textit{man} the boat'').
Similar considerations hold for \citet{wangHowCanLarge2024,wang-etal-2025-extracting} and \citet{Alternative:Sampling:giulianelli} that treat syntax as a deterministic function when a full sentence is known. 


\paragraph{Bayesian Approaches to  Sentence Processing}
Our proposal of capturing processing difficulty through belief update is conceptually similar to various Bayesian approaches to sentence processing. \citet{narayananBayesianModelsHuman2022, narayananBayesianModelPredicts2002} relate the processing difficulty of a word to the the belief reranking it induces over a set of parses calculated through a Bayes net.
However, this model is insensitive to changes to the probability distribution that do not trigger change in ranking.


This Bayesian view is also shared by predictive coding models of sentence processing, such as \citet{shainFMRIRevealsLanguagespecific2020}, which used surprisal as an analog of predictive error. More recent work by \citet{resnik:predictive:coding} provides a fully-specified model of predictive coding at the algorithmic level (in the sense of \citet{marr82}), specifically capturing belief update under predictive coding by quantifying the update through representation changes or the number of algorithm operations. However, it works on a continuous non-hierarchical latent state rather than linguistically plausible representations, and is tied to a specific inference algorithm. 

\paragraph{CCG Supertag Surprisal} by \citet{arehalliSyntacticSurprisalNeural2022} is most related to our work. It uses an incremental probabilistic model over possible CCG supertags $x$ for word $w_i$ \citep{steedmanSyntacticProcess2000} and compares predicted distribution before ($q$) and after ($p$) observing $w_i$. While the authors refer to supertagging as ``almost parsing'' it is important to note that they are not predicting a full sequence of supertags but only the upcoming supertag, i.e. supertags of each position are conditionally independent of each other. A new word cannot revise or penalize the decisions made on supertags for earlier words which is detrimental for modeling garden path sentences.
They compute syntactic surprisal by taking the negative logarithm of the expected supertag probability of the prior model $q$ under a true distribution $p$.

\begin{align}\label{eq:synsurp}
    \operatorname{S_{syn}}(p; q)=  
    - \log \sum_{x} p(x)q(x) 
\end{align}

This can actually be related to our R\'enyi framework in an interesting way: this equation is a R\'enyi Cross-Entropy for $\alpha=2$ (see Appendix~\ref{sec:app:renyi}).
This equation is not a divergence and does not measure the degree of change in the distributions--if $p$ and $q$ are the same (i.e. no new information provided by $w_i$) it would still predict non-zero processing work.
In other words, this equation does not satisfy the \textbf{identity of indiscernibles} defined in Section~\ref{sec:our:proposal:syntactic:belief:update} which would be an intuitive requirement of a processing difficulty measure.



\section{Methods}\label{sec:methods}
\label{sec:methods}
 
\subsection{Training Data}
We train our dependency parser on a combined corpus of English treebanks annotated under the Surface-Syntactic Universal Dependencies annotation scheme (SUD, \citealp{gerdesSUDSurfaceSyntacticUniversal2018}). In particular, we combine the EWT \cite{ewt}, GUM \cite{gum}, and GUMReddit \cite{gumreddit} treebanks, as well as the English portions of the LinES \cite{lines}, and PUD \cite{pud} multilingual treebanks. These treebanks were selected to get a broad sample of data from different sources with minimal compromises on consistency \cite{zeldesAreUDTreebanks2023}.

\subsection{Baseline Metrics}
We compare our formulation of Syntactic Belief Update against two baseline metrics: LLM lexical surprisal, and syntactic supertag surprisal computed following \citet{arehalliSyntacticSurprisalNeural2022}

LLM word surprisals serve as a baseline for the explanatory power of lexically-driven prediction, while syntactic surprisal serves as a baseline for lexically independent syntactic processing without an explicit formulation of belief update. To estimate these, we take the same pretrained RoBERTa model used to finetune our parser and train two new models over the same sentences in the training set of our parser. For lexical surprisal, we retrofit the RoBERTa model with a causal mask and finetune it on a next-word prediction task. For syntactic surprisal, we again retrofit with a causal mask, and finetune it on a joint next-word prediction and supertagging task following \citet{arehalliSyntacticSurprisalNeural2022}. Syntactic surprisal is then calculated as described in Equation \ref{eq:synsurp}, marginalizing supertag predictions over the prediction of the next word.
We closely match the training regime of both these models to our parser. Precise details for these and other baselines are described in Appendix \ref{sec:app:otherbaselines}.

\subsection{Empirical Data}
We evaluate our SBU and baseline metrics on their ability to predict human reading time on the garden path portion of the Syntactic Ambiguity Processing Benchmark (SAP, \citealp{huangLargescaleBenchmarkYields2024}). This is a large-scale self-paced reading time dataset collected from 2000 participants reading from from a large corpus sentences covering a range of syntactic ambiguities, as well as from naturalistic ``filler'' sentences designed to capture normal reading. 


The garden path portion of this corpus contains 24 sentences each of the three types of garden path constructions named above: NP/S, NP/Z, and MV/RR. Each sentence is matched with a disambiguated equivalent to serve as a control (See examples in (\ref{ex:nps}-\ref{ex:mvrr})). We thus operationalize the difficulty associated with a particular construction as the difference in reading time across the \textit{critical region}-- the three-word region after a sentence becomes disambiguated-- between the true garden path and its control counterpart.

\subsection{Statistical Analyses}\label{sec:r_analyses}
We fit two sets of linear mixed effects models to derive predictions about reading times over this critical region from Syntactic Belief Update metrics and baselines. First, we conduct a direct evaluation of the relationship between each metric and garden path processing by fitting the total reading time over the critical region directly to measures of that metric taken from all three words within the critical region, as well as nuisance predictors such as word length and word frequency. This allows us to assess the overall ability for our metric to capture garden path difficulty in isolation.

Then, we investigate the extent to which syntactic disambiguation during naturalistic reading might generalize to the specific reanalysis that human comprehenders are thought to undertake during garden path processing.  To do this, we fit linear mixed effects models with the same structure as before, but only over three-word regions taken during naturalistic reading. Success under this formulation would imply the stronger hypothesis that syntactic reanalysis is a special case of the syntactic disambiguation that engages during general reading. 

To compare how well the predicted reading times from these linear models align with the slowdowns shown by humans, we fit both predicted and empirical reading times using Bayesian mixed effects models that capture both item-level and construction-level slowdowns, following \citet{huangLargescaleBenchmarkYields2024}. 
Precise details for both our linear and Bayesian mixed effects models are described in Appendix \ref{sec:app:stats}.

\section{Results}\label{sec:results}
\label{sec:results}

\subsection{Syntactic Belief Update Uniquely Captures Slowdown Patterns Across Different Garden Path Types}
\begin{figure*}[tb]
    \begin{subfigure}{\textwidth}
        \includegraphics[width=0.95\linewidth]{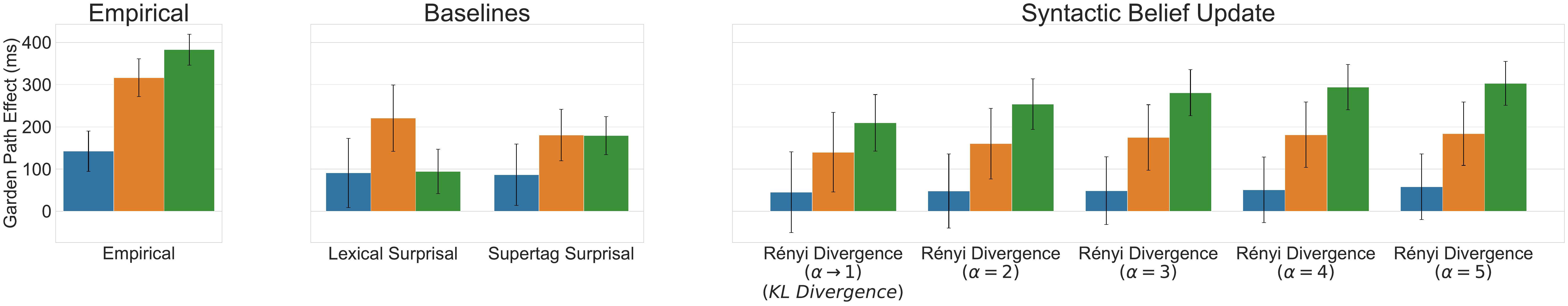}
        \caption{Fit on Critical Regions Directly}\label{fig:eois_roi}
    \end{subfigure}
    \begin{subfigure}{\textwidth}
        \includegraphics[width=0.95\linewidth]{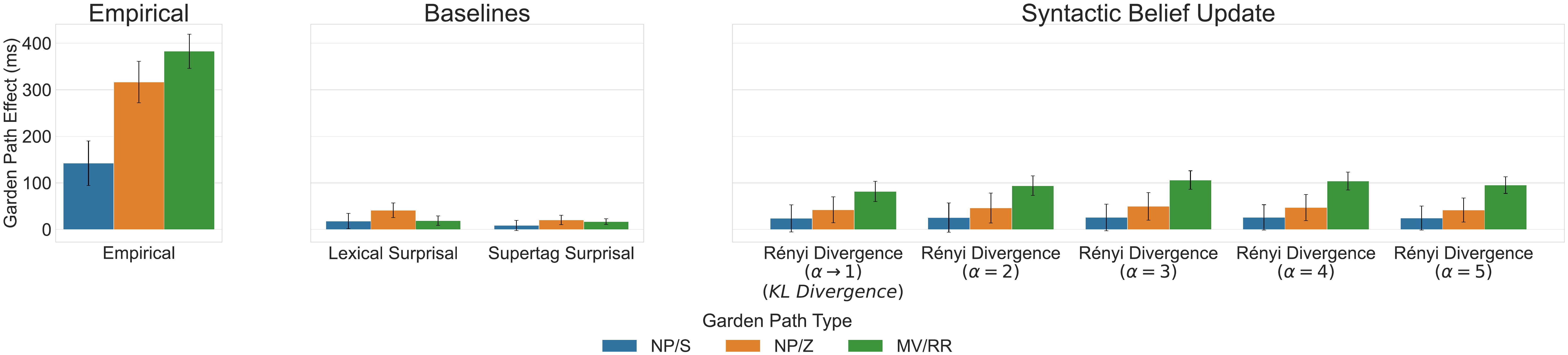}
        \caption{Fit on Naturalistic Filler Sentences Only}\label{fig:eois_filler}
    \end{subfigure}
    \caption{Estimated empirical garden path effects-- defined as the total slowdown over the critical region between a garden path and its corresponding control-- plotted against garden path effects estimated from predicted reading times fit over critical regions directly (a) and over naturalistic sentences only (b). Error bars show the 95\% CI.}\label{fig:eois}
\end{figure*}

We first validate the relationship between garden path reading time and our metrics of interest by fitting linear mixed effects models directly on critical region reading times. The aggregate predicted slowdowns derived from these models are shown in Figure~\ref{fig:eois_roi}, alongside the empirical equivalents taken from the SAP Benchmark.
We see that all metrics underpredict the absolute  difficulty of each garden path type by some degree.
However, only SBU is able to successfully capture the NP/S < NP/Z < MV/RR asymmetry. By contrast, lexical surprisal incorrectly predicts that the hardest MV/RR is on par with the easiest NP/S construction. Supertag surprisal does correctly predict NP/S to be the easiest construction, but fails to distinguish between NP/Z and MV/RR. We further note that increasing $\alpha$  for R\'enyi Divergence brings the predicted magnitude of slowdowns closer to the human range, suggesting human reanalysis difficulty is most sensitive to difference in the peaks of distributions covering a smaller number of plausible parses.

\begin{figure*}[tb]
    \begin{subfigure}{\textwidth}
        \includegraphics[width=0.95\linewidth]{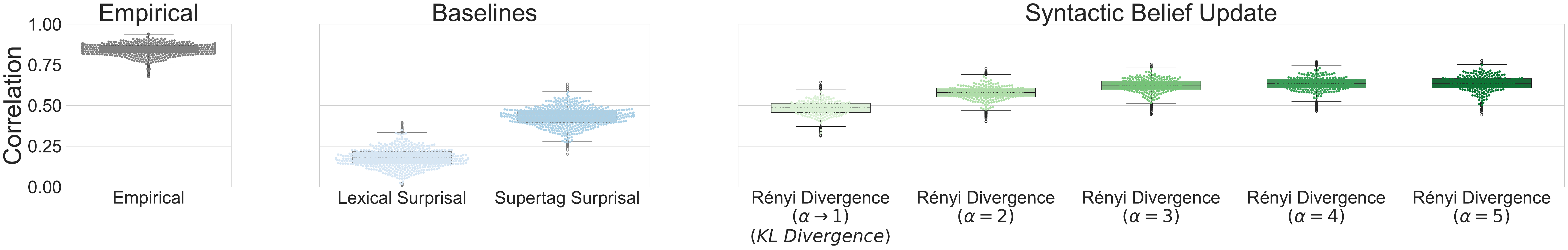}
        \caption{Fit on Critical Regions Directly}\label{fig:agg_corr_roi}
    \end{subfigure}
    \begin{subfigure}{\textwidth}
        \includegraphics[width=0.95\linewidth]{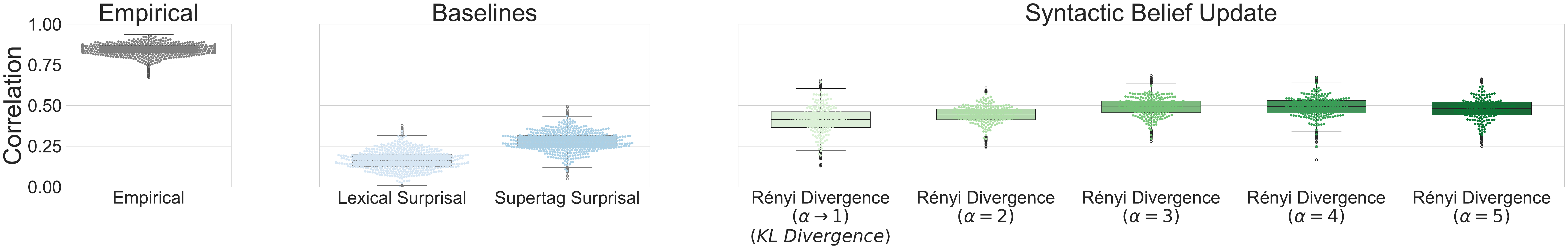}
        \caption{Fit on Naturalistic Filler Sentences Only}\label{fig:agg_cor_filler}
    \end{subfigure}
    \begin{subfigure}{\textwidth}
        \begin{subfigure}{0.41\textwidth}
            \includegraphics[width=0.95\linewidth]{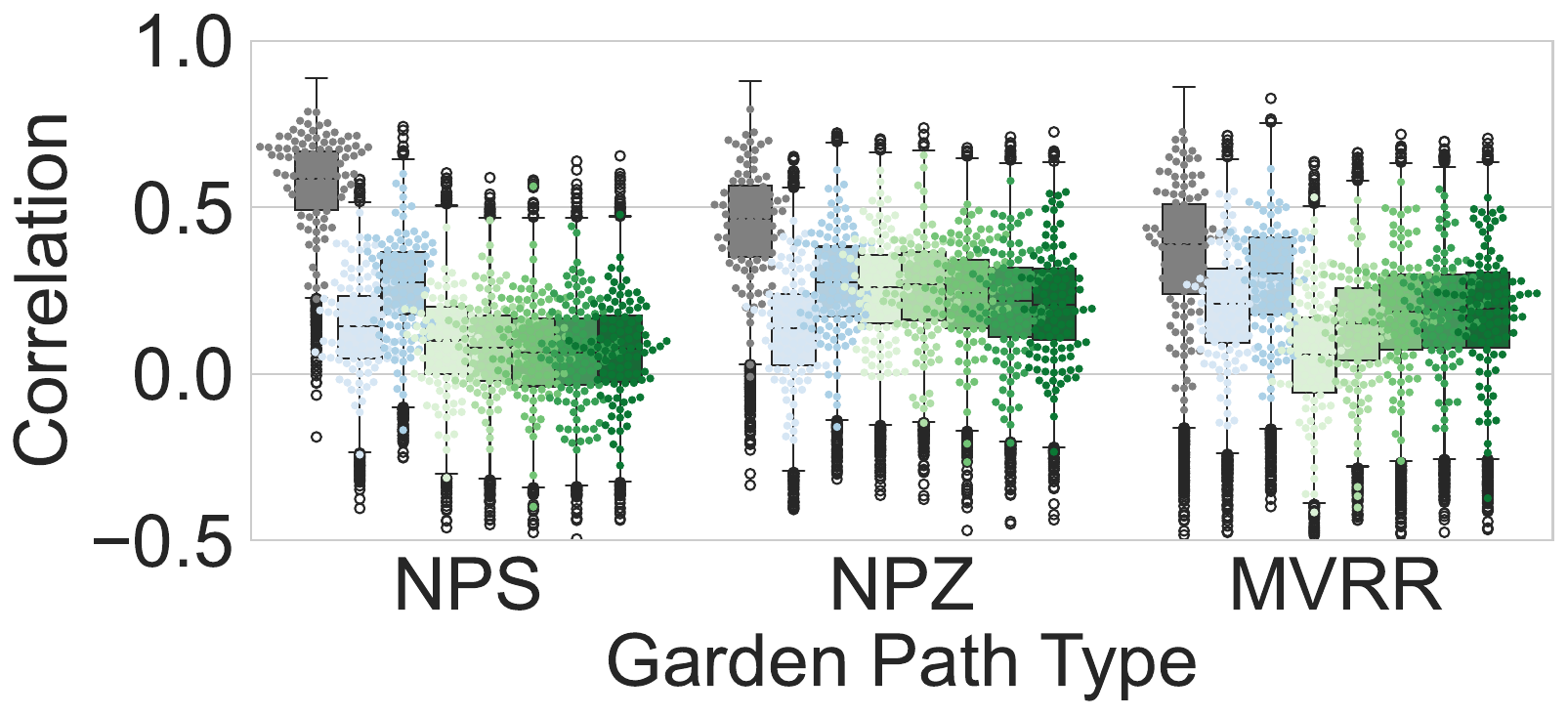}
            \caption{Fit on Critical Regions Directly}\label{fig:item_corr_roi}
        \end{subfigure}
        \begin{subfigure}{0.41\textwidth}
            \includegraphics[width=0.95\linewidth]{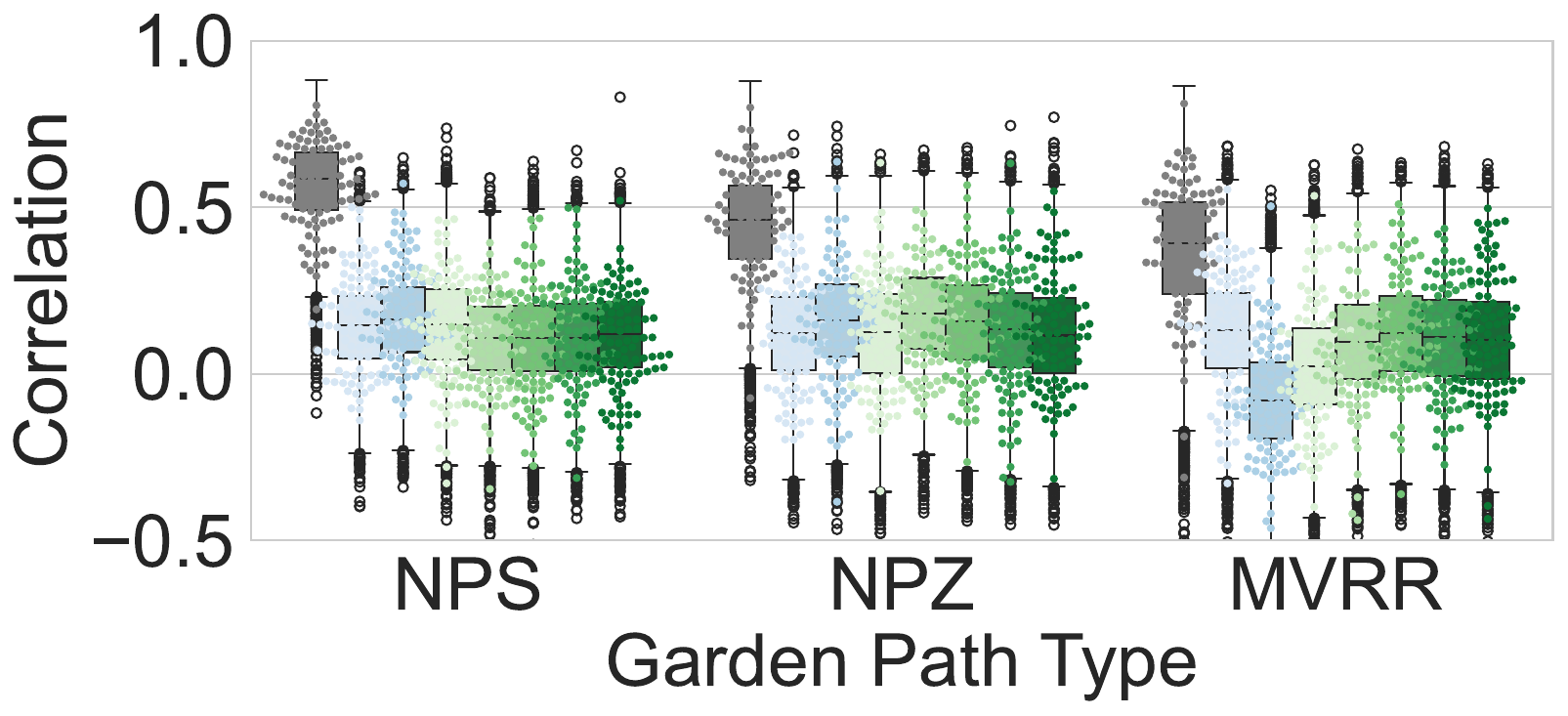}
            \caption{Fit on Naturalistic Filler Sentences Only}\label{fig:item_corr_filler}
        \end{subfigure}
        \includegraphics[width=0.17\linewidth]{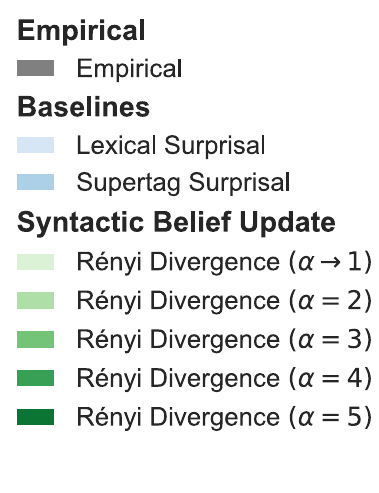}
    \end{subfigure}
    \caption{Sampled correlations between predicted slowdowns and and empirical slowdowns estimated from using a Monte Carlo process over our Bayesian mixed effects models, compared to the noise ceiling indexed by the sampled self-correlation over empirical slowdowns (grey). Panels (a) and (b) show the correlations over all items, while (c) and (d) show correlations within each garden path type. Box plots show the median and interquartile range; points beyond 1.5 times the interquartile range are shown as outliers.} 
    \label{fig:corrs}
\end{figure*}

To quantify how well the predicted slowdowns match the empirical slowdowns, we estimate the correlation between the predicted and empirical slowdowns by sampling from the itemwise posteriors obtained from our Bayesian mixed effects models (Figure \ref{fig:agg_corr_roi}). Here, we see some benefit for general lexically-independent syntactic prediction, with supertag surprisal showing much higher correlations than simple lexical surprisal. However, SBU shows higher correlations than both baselines, especially as $\alpha$ increases. We interpret these results in favor of not just syntactically-driven prediction, but \textit{specifically the update of prior syntactic beliefs}, as the strongest predictor of garden path difficulty.

\subsection{Syntactic Belief Update does not Capture Slowdowns \textit{Within} Construction}
We next ask if SBU can explain the item-level variance across garden paths within the same type. In other words, within a construction-type, is SBU predictive of which sentences are harder to process than others? To assess, we once again estimate item-level correlations by sampling from our itemwise posteriors, but compute correlations within each garden path type (Figure \ref{fig:item_corr_roi}).

We first note far greater noise in these within-construction correlations, with far lower noise ceilings as indexed by the empirical self-correlations, and a high degree of overlap across all metrics. We do observe however, that the slowdowns predicted by each metric are generally positively correlated, and that supertag surprisal generally shows higher correlations than all other metrics, including SBU. These results seem to suggest that syntactic predictability, rather than Syntactic Belief Update, may explain more of the within-construction slowdowns once garden path type is controlled for.


\subsection{Syntactic Belief Update Partially Extends Naturalistic Reading Times to Garden Path Difficulty}
Finally, we examine if the linking between syntactic belief and processing difficulty over naturalistic sentences will also explain garden path reanalysis. To do this we fit the same linear models predicting reading time over all three-word regions in the naturalistic filler portion of the SAP dataset, and assess their ablitity to predict garden path difficulty in terms of magnitude (Figure \ref{fig:eois_filler}) and correlation (Figure (\ref{fig:agg_cor_filler}, \ref{fig:item_corr_filler})).
Success would imply that SBU indexes some processing mechanism that engages for both naturalistic sentences and garden paths.

We find that overall magnitudes and correlations with empirical slowdowns both suffer under this approach. However, SBU predicts  higher magnitudes over garden paths and maintains much higher overall correlation than both lexical and supertag baselines. SBU also remains the only metric to capture the difficulty hierarchy across the different garden path types, while matching the other metrics in terms of its ability to capture item-level correlations within each type. 

These results suggest that SBU captures some commonality between naturalistic syntactic disambiguation and garden path processing in terms of cross-construction asymmetries. However, while SBU captures more of the garden path magnitude than previous proposals, it still does not capture the full magnitude reported in \citet{huangLargescaleBenchmarkYields2024}.

\section{Discussion and Conclusion}\label{sec:discussion}

Compared to all our baseline metrics, Syntactic Belief Update uniquely captures the cross-construction difficulty patterns that human readers experienced across garden paths, and predicts a larger garden path effect even when fit only on naturalistic reading times. As our formulation of Syntactic Belief Update isolates the difficulty of the update to syntactic beliefs at the garden path critical point, we take these results in favor of this belief update as the main driver of garden path processing difficulty across constructions.

However, we find that SBU does not explain item-level variations in reading time once garden path type is controlled for. To understand this finding, we note that SBU depends primarily on updates over the structural distributions within the critical region. However, in the controlled garden path set that we test on, structural distributions before and after the critical region are nearly identical \textit{within a single garden path type}. While the lexical identity of the words within and before the critical region do influence syntactic beliefs to some degree, our results suggest that this influence is no more predictive than baselines in explaining item-level variations within the same garden path type. 

We also note stronger fits to human performance when computing syntactic belief update using R\'enyi Divergences with higher alpha. This result shows how the choice of divergence metric informs conclusions about human sentence processing. In this case, the stronger fit to higher alpha suggests that the human sentence processor is more sensitive to shifts to beliefs about the most likely parses than it is to minor adjustments in the long tail of unlikely parses. One explanation for this finding comes from \textit{limited-parallelism} models of sentence processing in which less likely parses are simply not maintained, e.g. in particle filter \citep{levyModelingEffectsMemory2008} and beam search parsing strategies \citep{berta:franzluebbers-etal-2024-multipath}.  

Finally, while the focus of this paper has been on garden path sentences specifically, our framework is in principle applicable to sentence processing more broadly. We plan to apply our framework to additional processing puzzles such as relative clause asymmetries and center-embedding, in which alternative divergence metrics that factor in the effect of memory on belief update may be necessary to explain human performance \citep{resnikLeftcornerParsingPsychological1992, gibsonDependencyLocalityTheory2000}. We also plan to apply our framework to general processing over typical sentences. These analyses will help elucidate whether the reanalysis we observed in this study is unique to garden paths, or if it is simply a special case of some general mechanism that also engages during  sentence processing more broadly. 

\section*{Limitations}
\paragraph{Assumption of Linearity}
We note an implicit assumption of a linear relationship between belief update and processing difficulty by our use of a linear mixed effects model following standard practices in the surprisal literature \citep{smithEffectWordPredictability2013, wilcoxLanguageModelQuality2023, wilcoxTestingPredictionsSurprisal2023}. However, a existing body of literature have argued for that the relationship between belief update and reading time is in fact superlinear \cite{hooverPlausibilitySamplingAlgorithmic2023, viglyComprehensionEffortCost2025, meisterRevisitingUniformInformation2021}, on the basis of both empirical fit and algorithmic plausibility through sampling. We plan to further explore the precise nature of the linking relationship between belief update and processing difficulty in future work. In particular, a super-linear relationship between belief update and processing difficulty could explain why reading time prediction fit over typical naturalistic sentences underpredict garden path reading time.

\paragraph{Choice of Metric and Representation}
Our proposal of Syntactic Belief Update is intended as a general framework for quantifying syntactic reanalysis, compatible with any combination of metric and syntactic formalism. This work only presents results derived from one particular combination: the R\'enyi Divergences over Surface-Syntactic Universal Dependency Trees. Preliminary results showed that this choice is not trivial-- we found that standard Universal Dependencies trees and metrics derived from cross-entropy often derived results that do not match the human pattern. Though we find R\'enyi Divergence and SUD trees to be reasonable initial hypotheses, it very well may be that human sentence processing is better explained by other formalisms and metrics, which we leave open for future work.

\paragraph{Empirical Coverage}
We test our model on only English, and only over three different types of garden paths. While these are the garden path types with the most empirical data, and which  have been most extensively studied by psycholinguistics, they are are far from the only garden path ambiguities to exist (for a survey just in English, see \citealp{lewisArchitecturallybasedTheoryHuman1993a}). Garden paths that appear in languages other than English also possess a number of interesting properties that do not manifest in English, from interactions with morphology \cite{peristeriEffectsExecutiveAttention2020} and syntactic processes such as scrambling \cite{baderSubjectObjectAmbiguitiesGerman1999}. While the general principles of SBU should hold for these other types of garden paths, whether or not our results will extend to these other types of garden paths remains an empirical question.

\paragraph{Timecourse of Garden Path Processing}
In the experiments presented here, we were primarily interested in capturing the general difficulty of each garden path type. Thus, we operationalized garden path difficulty as the total slowdown that readers experience over the critical region. However, there are also interesting patterns within the critical region that we do not model. In particular, \citet{huangLargescaleBenchmarkYields2024} show that NP/Z garden paths show a larger slowdown in the critical region far sooner than the other two garden path types. We hypothesize that these differences are due to interactions between reanalysis and spillover that our model does not currently account for. 

\section*{Acknowledgments}
We would like to thank the members of the Hale Lab and the Group for Language and Intelligence for their comments and feedback. We would especially like to thank Donald Dunagan for helpful discussions regarding analyses and the setup of statistical models.

\bibliography{custom, misc}

\appendix

\section{R\'enyi Entropy, Cross-Entropy and Divergence for Markov Random Fields}
\label{sec:app:renyi}

\citet{renyi-entropy-original} has introduced a generalization of Shannon's entropy and Kullback-Leibler (KL) divergence. This generalization keeps some of the most important properties, e.g. additivity of independent events, while being able to cover many special cases with a single $\alpha$ parameter. For example, for $\alpha \rightarrow 0$ we have Hartley entropy, for $\alpha \rightarrow 1$ we have Shannon's entropy, for $\alpha=2$ we get collision entropy, for $\alpha \rightarrow \infty$ we get min-entropy. Similar generalizations hold for R\'enyi's generalization of divergence. R\'enyi Cross-Entropy was proposed in follow-up works by different researchers who found different generalizations. Here we present the most prominent generalization by \citet{valverde-albacetee-pelaez-moreno} in a simplified form from \citet{Thierrine-2022-entropy}.

Below are the equations that define R\'enyi Entropy (\ref{eq:a:renyi:entropy}), Cross-Entropy (\ref{eq:a:renyi:cross:entropy}) and Divergence (\ref{eq:a:renyi:divergence}):
\begin{align}
    H&_{\alpha}(p) = \frac{1}{1-\alpha} \log \sum_{x} p(x)^\alpha \label{eq:a:renyi:entropy} \\
    H&_{\alpha}(p; q) = \frac{1}{1-\alpha} \log \sum_{x} p(x)q(x)^{\alpha-1} \label{eq:a:renyi:cross:entropy}\\
    D&_{\alpha}(p \mid\mid q) = \frac{1}{\alpha-1}\log \sum_{x}p(x) \left(\frac{p(x)}{q(x)}\right)^{\alpha-1} \label{eq:a:renyi:divergence}
\end{align}

All the equations above specify how to compute the quantities by enumerating all possible events but very often that is not possible because the event space is exponentially large. This is specifically the case in our work where the number of single-root edge dependency trees grows super-exponentially -- for $n$ words there are $n^{n-1}$ single-root dependency trees \citep{stanojevic-2022-unbiased}. In those cases it is necessary to have a compact representation of events using graphical models that will exploit internal shared structure to compute sufficient statistics efficiently. As it turns out, the only statistic we need is a partition function which luckily we can compute efficiently for edge-factored model over dependency trees using Matrix-Tree Theorem \citep{kooStructuredPredictionModels2007,smith-smith-2007-probabilistic,mcdonald-satta-2007-complexity,zmigrodEfficientComputationExpectations2021}.

In sub-sections below we show the derivation of how each of the R\'enyi quantities can be reduced to the computation of partition functions. The quantities are computed between distributions $p$ and $q$ that we assume have the same graphical structure but not the same potentials assigned to the parts of the structure. All partition functions follow the notation $Z_{p^aq^b}$ where $p^a q^b$ represents distribution that we get by combining potentials of $p$ and $q$ with $\phi_{p^aq^b}(t)=\phi_p(t)^a \phi_q(t)^b$. Because $t$ factorizes the same in both $p$ and $q$ we can get combined distribution by just locally multiplying potentials of each edge $\phi_{p^aq^b}(e)=\phi_p(e)^a \phi_q(e)^b$. It is also important to distinguish $Z_{p^a}$ from $Z_p^a$ -- the first one applies power $a$ to potentials, while the second one applies it to the partition function.

\subsection{R\'enyi Entropy}
\label{sec:app:renyi:entropy}

\begin{align}
    H&_{\alpha}(p) \nonumber \\
    & = \frac{1}{1-\alpha} \log \sum_{x} p(x)^\alpha \nonumber \\
    & = \frac{1}{1-\alpha} \log \sum_{x} \frac{\phi_p(x)^\alpha}{Z_p^\alpha} \nonumber \\
    & = \frac{1}{1-\alpha} \log \frac{1}{Z_p^\alpha}\sum_{x} \phi_p(x)^\alpha  \nonumber \\
    & = \frac{\log Z_{p^\alpha} - \alpha \log Z_p}{1-\alpha}
\end{align}

\subsection{R\'enyi Cross-Entropy}
\label{sec:app:renyi:cross:entropy}

\begin{align}
    H&_{\alpha}(p; q) \nonumber \\
    & = \frac{1}{1-\alpha} \log \sum_{x} p(x)q(x)^{\alpha-1} \nonumber\\
    & = \frac{1}{1-\alpha} \log \sum_{x} \frac{\phi_{p}(x)}{Z_p}\frac{\phi_{q}(x)^{\alpha-1}}{Z_q^{\alpha-1}} \nonumber\\
    & = \frac{1}{1-\alpha} \log  \frac{1}{Z_p Z_q^{\alpha-1}} \sum_{x} \phi_{p}(x)\phi_{q}(x)^{\alpha-1} \nonumber\\
    & = \log Z_{q} + \frac{\log Z_{pq^{\alpha-1}} - \log Z_{p}}{1-\alpha} 
\end{align}

\subsection{R\'enyi Divergence}
\label{sec:app:renyi:divergence}

\begin{align}
    D&_{\alpha}(p \mid\mid q) \nonumber \\
    & = \frac{1}{\alpha-1}\log \sum_{x}p(x) \left(\frac{p(x)}{q(x)}\right)^{\alpha-1} \nonumber \\
    & = \frac{1}{\alpha-1}\log \sum_{x}p(x)^{\alpha} q(x)^{1-\alpha} \nonumber \\
    & = \frac{1}{\alpha-1} \log \sum_{x} \frac{\phi_{p}(x)^{\alpha}}{Z_p^{\alpha}}\frac{\phi_{q}(x)^{1-\alpha}}{Z_q^{1-\alpha}} \nonumber\\
    & = \frac{1}{\alpha-1} \log  \frac{1}{Z_{p}^{\alpha}Z_{q}^{1-\alpha}} \sum_{x} \phi_{p}(x)^{\alpha}\phi_{q}(x)^{1-\alpha} \nonumber\\
    & = \log Z_{q} + \frac{\log Z_{p^{\alpha}q^{1-\alpha}} - \alpha \log Z_{p}}{\alpha-1}
\end{align}

\subsection{R\'enyi for $\alpha \rightarrow 1$}

As mentioned before R\'enyi (Cross-)Entropy and Divergence for $\alpha$ approaching $1$ reduce to Shannon (Cross-)Entropy and KL Divergence. However, we cannot use equations above to compute those quantities because of numerical instability when $\alpha$ gets close to $1$. For this special case of $\alpha \rightarrow 1$ we use equations from \citet{stanojevicSynJaxStructuredProbability2023}:
\begin{align}
    &H_{\alpha\rightarrow 1}(p; q) = \log Z_q - \sum_{e \in E} p(e) \phi_q(e) \\
    &H_{\alpha\rightarrow 1}(p) = H_{\alpha\rightarrow 1}(p; p) \\
    &D_{\alpha\rightarrow 1}(p \mid\mid q) = H_{\alpha\rightarrow 1}(p; q) - H_{\alpha\rightarrow 1}(p)
\end{align}

These equations hold only for $\alpha\rightarrow 1$.

\section{Details for Model Training}\label{sec:app:training}
\subsection{Parser}
To get the potentials of each edge $\phi(i, j)$, we finetune a bi-affine attention layer \cite{dozatDeepBiaffineAttention2017} on top of a pretrained RoBERTa model (\texttt{roberta-large}; \citealp{roberta}) to maximize Equation \ref{eq:tree_prob}.
For each word in a tree, we obtain encoding $e_i$ by taking the mean representation across its subtokens in the final layer of RoBERTa.
Then, $\phi(i, j)$ is computed as:
\begin{align}
    &h_{i} = \text{MLP}_{head}\left(e_i \right),\ h_{j} = \text{MLP}_{dep}\left(e_j \right)\label{mlp}\\
    &\phi(i, j) =  \exp(h_{i}^\top \textbf{W} h_{j} + \textbf{u}^\top h_{i} + \textbf{v}^\top h_j + \text{b})
\end{align}

We alternate between two different objectives during training. The first objective, in place for the first 1000 training steps, is a local loss that treats each word's head prediction as an independent classification task. The second objective, in place for the rest of training, then optimizes directly the log probability of the given tree given in Equantion \ref{eq:tree_prob}. As a large portion of parsing decisions can be attributed to relatively local preferences, we intend the local objective to encourage the model to stabilize to a reasonable solution, before utilizing the full log probability to integrate the consideration of more global structure. 

Partial prefix trees are presented to the parser 90\% of the time, obtained by uniformly sampling an index for each tree and masking out the lexical identity of all nodes past that index. Each MLP is a one-layer fully-connected layer that preserves the 1024 dimensionality of the RoBERTa embeddings followed by ReLU. We train our parser for a total of 50 epochs with an effective batch size of 256 using the Adam optimizer with initial learning rate 1e-5. Our parser converged at 10 epochs with an overall Unlabeled Attachment Score of 0.78 over both full and prefix trees (0.94 over just full trees).

\subsection{Baselines}
To estimate our baseline metrics of lexical surprisal and supertag surprisal, we take a RoBERTa model with the same architecture as above and retrofit it with a causal mask. We then finetune one instance of this causal RoBERTa on a next-token prediction task to compute lexical surprisal, and another on a joint next-word prediction and super-tag prediction task for syntactic surprisal following \cite{arehalliSyntacticSurprisalNeural2022}. All tasks are accomplished with an MLP with the exact same architecture as the head and dep-scoring MLPs in Equation~\ref{mlp}. Similar to our parser, we train our baseline models for 50 epochs with an effective batch size of 256, using the Adam optimizer with initial learning rate 1e-5, and selecting the epoch with the best validation loss for evaluation. Our causal RoBERTa model converged in 11 epochs with a token perplexity of 74.14, while the joint syntactic surprisal model converged in 28 epochs with a next-word prediction accuracy of 0.22 and a supertag prediction accuracy of 0.86.

\section{Details for Statistical Analyses}\label{sec:app:stats}
\subsection{Deriving Reading Time Predictions with Linear Mixed Effects}\label{sec:app:lmer}
To directly estimate the relationship between a metric and critical region reading time, we fit linear mixed effects models predicting the total reading time over the critical region across ambiguous and disambiguated versions of each garden path sentence. Each model contains fixed effects for the metric of interest at each word within the critical region, as well as fixed baseline terms consisting of the length and unigram length frequency interaction over the same words. We also include a per-item random intercept and per-participant random slopes and intercept. Expressed in Wilkinson-Rogers notation \citep{wilkinson:rogers:1973}:

\begin{align*}
    \text{RT}&_{critical} \sim \text{metric}_{w_0} + \text{metric}_{w_1} + \text{metric}_{w_2} \\
                 &+ \text{Length}_{w_0} * \text{LogFreq}_{w_0}\\
                 &+ \text{Length}_{w_1} * \text{LogFreq}_{w_1}\\
                 &+ \text{Length}_{w_3} * \text{LogFreq}_{w_2}\\
                 &+ (1\!+\!\text{metric}_{w_0}\!+\!\text{metric}_{w_1}\\
                 &\quad\quad+\!\text{metric}_{w_2} \|\ \text{participant}) \\
                 &+ (1\ | \text{ item})
\end{align*}
Note we use $x_{w_i}$ to denote a metric taken at the $i$th word of the critical region, and that mixed effect models are written in standard R notation, where interaction terms such as $\text{Length} \ * \ \text{Frequency}$ are implicitly expanded as $\text{Length}\ +\ \text{Frequency} \ +\ \text{Length} \ \times \ \text{Frequency}$. Fitting was done in R using the \texttt{lme4} package \cite{lmer}, using the BOBYQA algorithm with 200,000 maximum evaluations.

To estimate how well reading time predictions derived from non-garden-path sentences might generalize to unseen garden paths, we fit models with the exact same structure, but over a set of naturalistic filler sentences that do not appear in the garden path set. As there is no longer a single ``critical region'' to consider, we fit now the overall reading time for every three-word block, using the same effect structure over metrics derived from the individual words within that block. 
\begin{align}
    \text{RT}&_{total} \sim \text{metric}_{w_0} + \text{metric}_{w_1} + \text{metric}_{w_2} \nonumber\\
                              &+ \text{Length}_{w_0} * \text{LogFreq}_{w_0} \nonumber\\
                              &+ \text{Length}_{w_1} * \text{LogFreq}_{w_1} \nonumber\\
                              &+ \text{Length}_{w_3} * \text{LogFreq}_{w_2} \nonumber\\
                              &+ (1\!+\!\text{metric}_{w_0}\!+\!\text{metric}_{w_1}\nonumber\\
                              &\quad\quad+\!\text{metric}_{w_2} \|\ \text{participant})\nonumber \\
                              &+ (1\ | \text{ item}) \nonumber
\end{align}

\subsection{Comparing Predicted and Empirical Garden Path Difficulty with Bayesian Mixed Effects}\label{sec:app:brms}
We quantify the difficulty associated with a particular garden path sentence as the expected slowdown in reading time between an ambiguous garden path and its disambiguated control over the critical region. To estimate this slowdown, we fit a Bayesian mixed effects model using garden path type and the prescence of an ambiguity as binary effects. In particular, given a set of total RTs over a set of ambiguous and disambiguous garden paths, we fit the model:
\begin{align*}
    \text{RT}_{critical} &\sim \text{IsAmbiguous} * \text{GPType} \\
                    &+ (1 + \text{IsAmbiguous} * \text{GPType }  \| \text{ item}) \\
                    &+ (1 \ |\  \text{ participant}) \\
\end{align*}
This model is fit once over the empirical critical RT, and then once for each set of predicted RTs obtained from the linear models described above using the \texttt{brms} package \cite{brm1, brm2, brm3}. Each model is fit with four sampling chains of 6,000 iterations, a maximum tree depth of 10, and an acceptance probability of 0.8 each, discarding the first 3,000 iterations as warm-up samples. If any parameter estimate of a model this converged a $\hat{R} > 1.05$, we restart the model with increased iterations until they do so, up to 12,000 iterations. If a model still did not fully converge, we simplified the random effect structure by removing the per-participant intercept. This was required for R\'enyi Divergence as $\alpha \rightarrow 1$ fit over naturalistic sentences, and supertag surprisal fit over critical regions only. 

This process yields a set of empirical and predicted garden path effects that can be compared on aggregate. Overall garden path effect are obtained by reading off the fixed-effect coefficients (e.g. IsAmbiguous + IsAmbiguous:NP/S to obtain the garden path effect for NP/S). Item-level effects of interest can be  similarly obtained using the item-level random effects for ambiguity and garden patht type. These models are fit in R using the \texttt{brms} package, adapting code from \citet{huangLargescaleBenchmarkYields2024}.

To further examine how well a metric captures item-level effects, we perform a Monte Carlo simulation to estimate the correlation between the predicted and empirical slowdowns over all items, following \citet{huangLargescaleBenchmarkYields2024}. For each garden path and control pair, we sample an empirical slowdown and a predicted slowdown from the posterior distribution of the estimated Bayesian model. We then take the Pearson correlation across these slowdowns across all items, and across each garden path type. These are compared to the self-correlation over the empirical items, repeating the same process by correlating across samples from the same estimated distribution of the empirical slowdowns. Figure \ref{fig:corrs} of the main text reports distributions over correlations sampled 10,000 times from this procedure.

\section{Other Metrics}
\subsection{Other Baselines} \label{sec:app:otherbaselines}
\begin{figure*}[tb]
    \begin{subfigure}{\textwidth}
        \includegraphics[width=\linewidth]{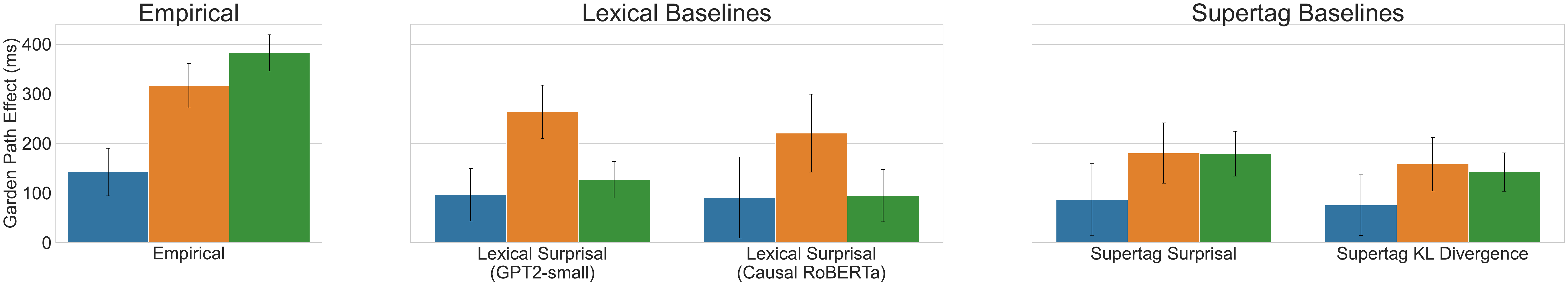}
        \caption{Fit on Critical Regions Directly}\label{fig:eois_roi_baseline}
    \end{subfigure}
    \begin{subfigure}{\textwidth}
        \includegraphics[width=\linewidth]{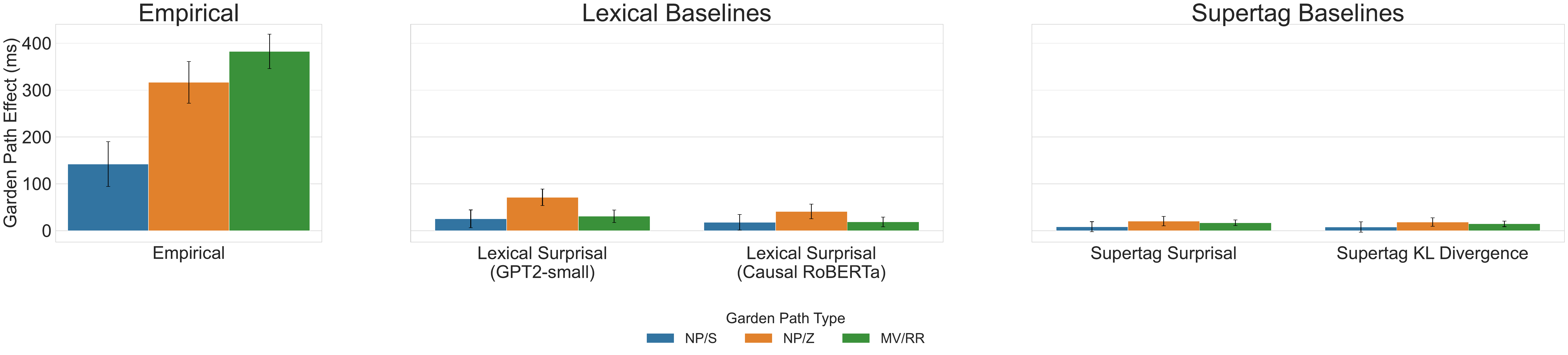}
        \caption{Fit on Naturalistic Filler Sentences Only}\label{fig:eois_filler_baseline}
    \end{subfigure}
    \caption{Comparison of empirical and predicted garden path effects from our alternative baselines, fit on critical regions directly (a) and over naturalistic sentences only (b). }\label{fig:baseline_eois}
\end{figure*}
\begin{figure*}[tb]
    \begin{subfigure}{\textwidth}
        \includegraphics[width=\linewidth]{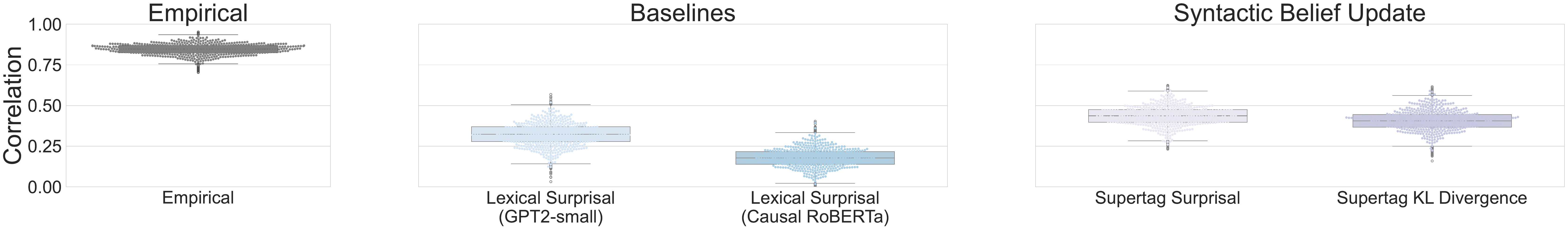}
        \caption{Fit on Critical Regions Directly}\label{fig:agg_corr_roi_baseline}
    \end{subfigure}
    \begin{subfigure}{\textwidth}
        \includegraphics[width=\linewidth]{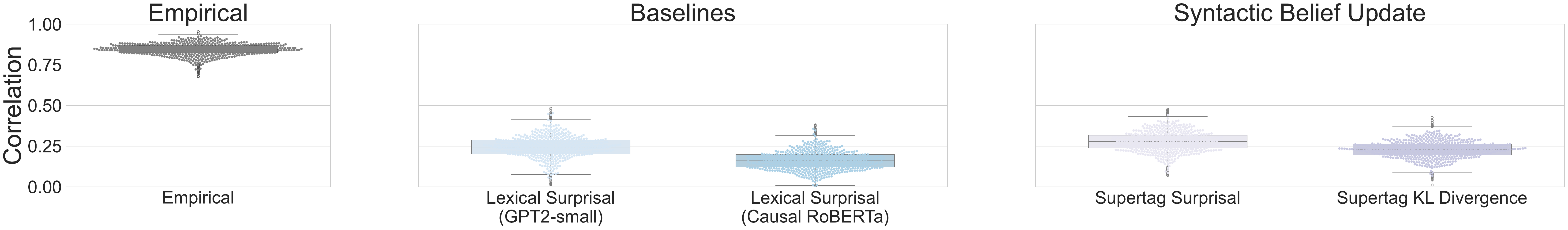}
        \caption{Fit on Naturalistic Filler Sentences Only}\label{fig:agg_cor_filler_baseline}
    \end{subfigure}
    \begin{subfigure}{\textwidth}
        \begin{subfigure}{0.41\textwidth}
            \includegraphics[width=\linewidth]{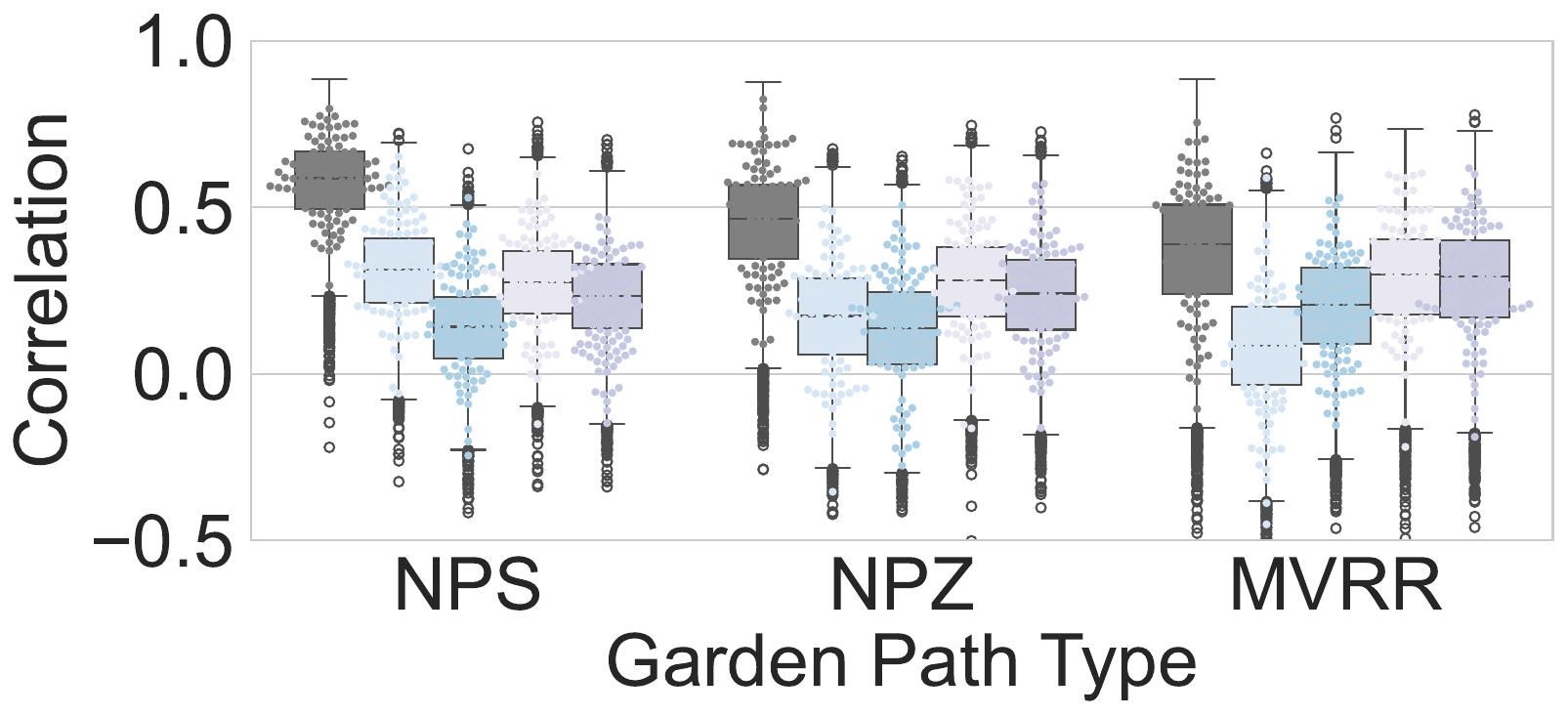}
            \caption{Fit on Critical Regions Directly}\label{fig:item_corr_roi_baseline}
        \end{subfigure}
        \begin{subfigure}{0.41\textwidth}
            \includegraphics[width=\linewidth]{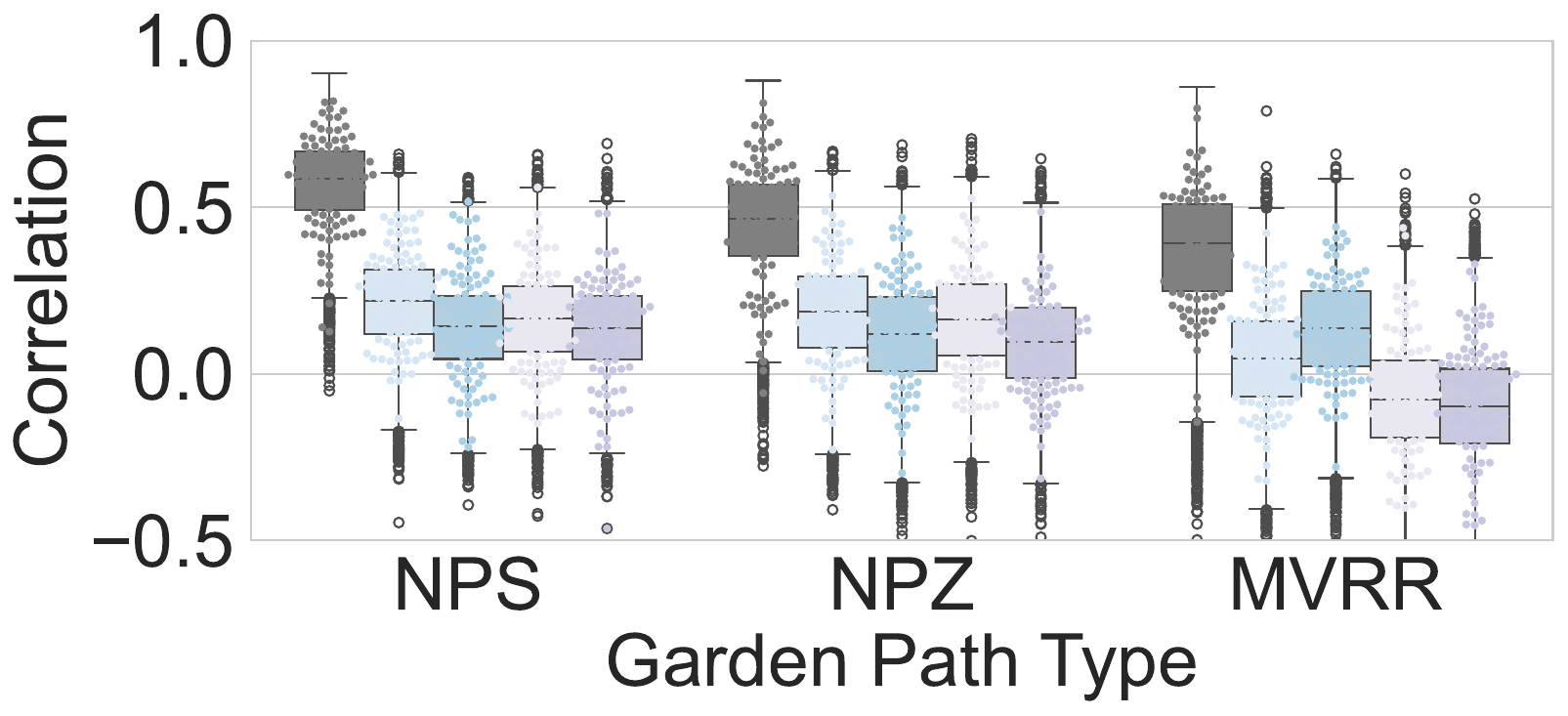}
            \caption{Fit on Naturalistic Filler Sentences Only}\label{fig:item_corr_filler_baseline}
        \end{subfigure}
        \includegraphics[width=0.17\linewidth]{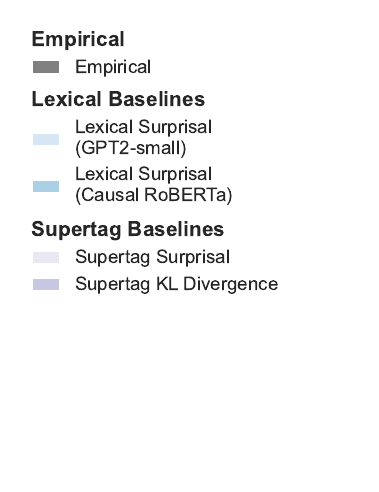}
    \end{subfigure}
    \caption{Sampled correlations between predicted slowdowns and and empirical slowdowns from our alternative baselines. Panels (a) and (b) show the correlations over all items, while (c) and (d) show correlations within each garden path type.} 
    \label{fig:corrs_sup}
\end{figure*}
In addition to the baseline metrics presented in the main paper, we also considered two other baselines that ended up yielding similar results: lexical surprisal extracted from GPT2-small \cite{radford2019language}, and supertag surprisal computed using KL divergence rather than R\'enyi Quadratic Cross-Entropy. We considered GPT2-small surprisal due to its widespread use as a standard psycholinguistic model for surprisal \cite{ohWhyDoesSurprisal2023}, and we considered KL divergence to test if metric satisfying the identity of indiscernibles might yield more human-like results over supertags. We found few differences between the lexical surprisals taken from GPT2-small and our causal RoBERTa, and between KL divergence and R\'enyi Quadratic Cross-Entropy over supertags, and so omitted from our main results (Figures \ref{fig:baseline_eois} and \ref{fig:corrs_sup}).

\subsection{R\'enyi Cross-Entropies over Parser Belief Distributions}
\begin{figure*}[tb]
    \begin{subfigure}{\textwidth}
        \includegraphics[width=0.98\linewidth]{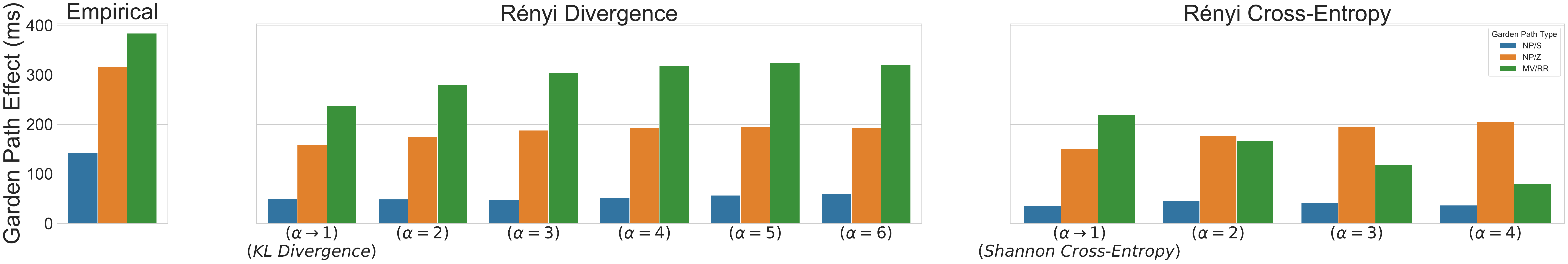}
        \caption{Empirical and Predicted Garden Path Effects }
    \end{subfigure}
    \begin{subfigure}{\textwidth}
        \includegraphics[width=0.98\linewidth]{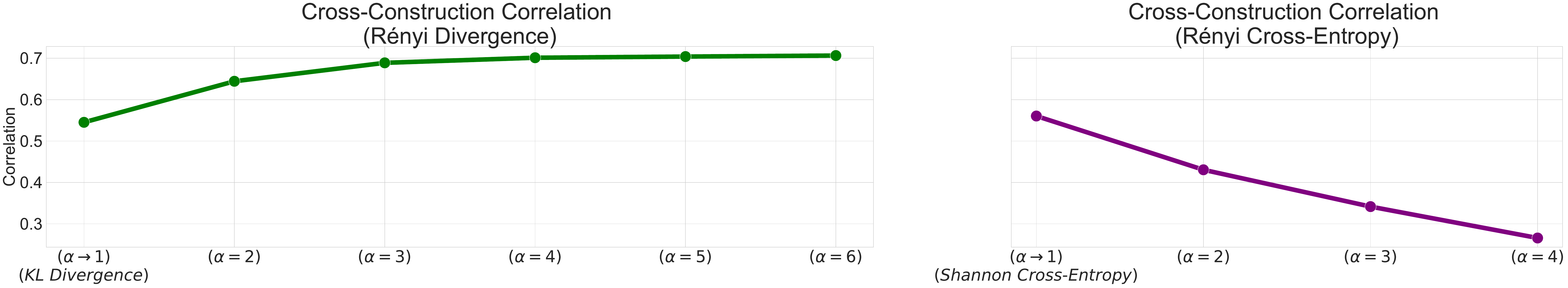}
        \caption{Correlation with Empirical}
    \end{subfigure}
    \caption{Frequentist comparison of R\'enyi Divergence and R\'enyi Cross-Entropy for different values of $\alpha$}\label{fig:app:alphasweep}
\end{figure*}
\begin{figure*}[tb]
    \begin{subfigure}{\textwidth}
        \begin{subfigure}{0.48\linewidth}
            \includegraphics[width=\linewidth]{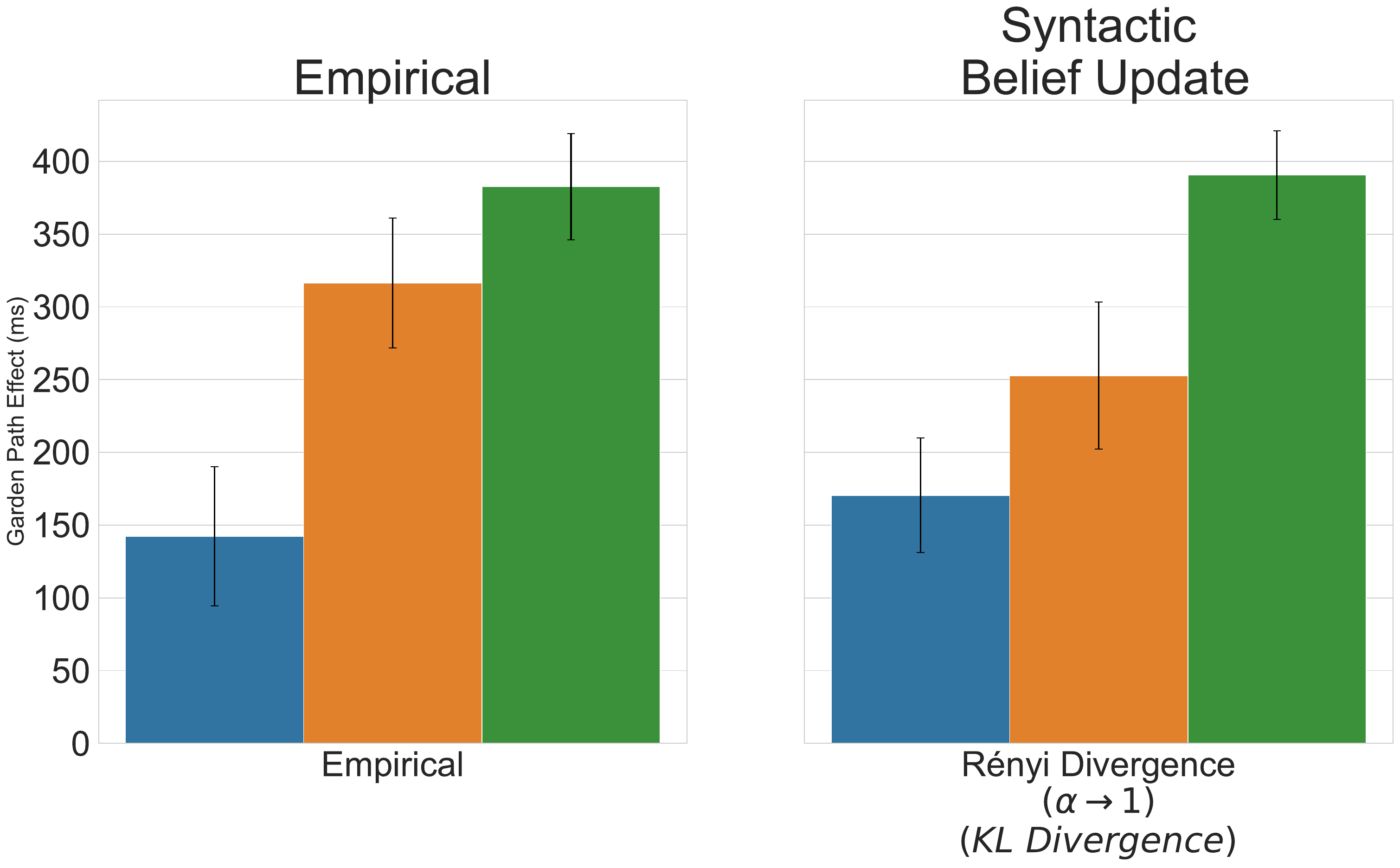}
            \caption{Empirical and Predicted Garden Path Effect}
        \end{subfigure}
        \begin{subfigure}{0.48\linewidth}
            \includegraphics[width=\linewidth]{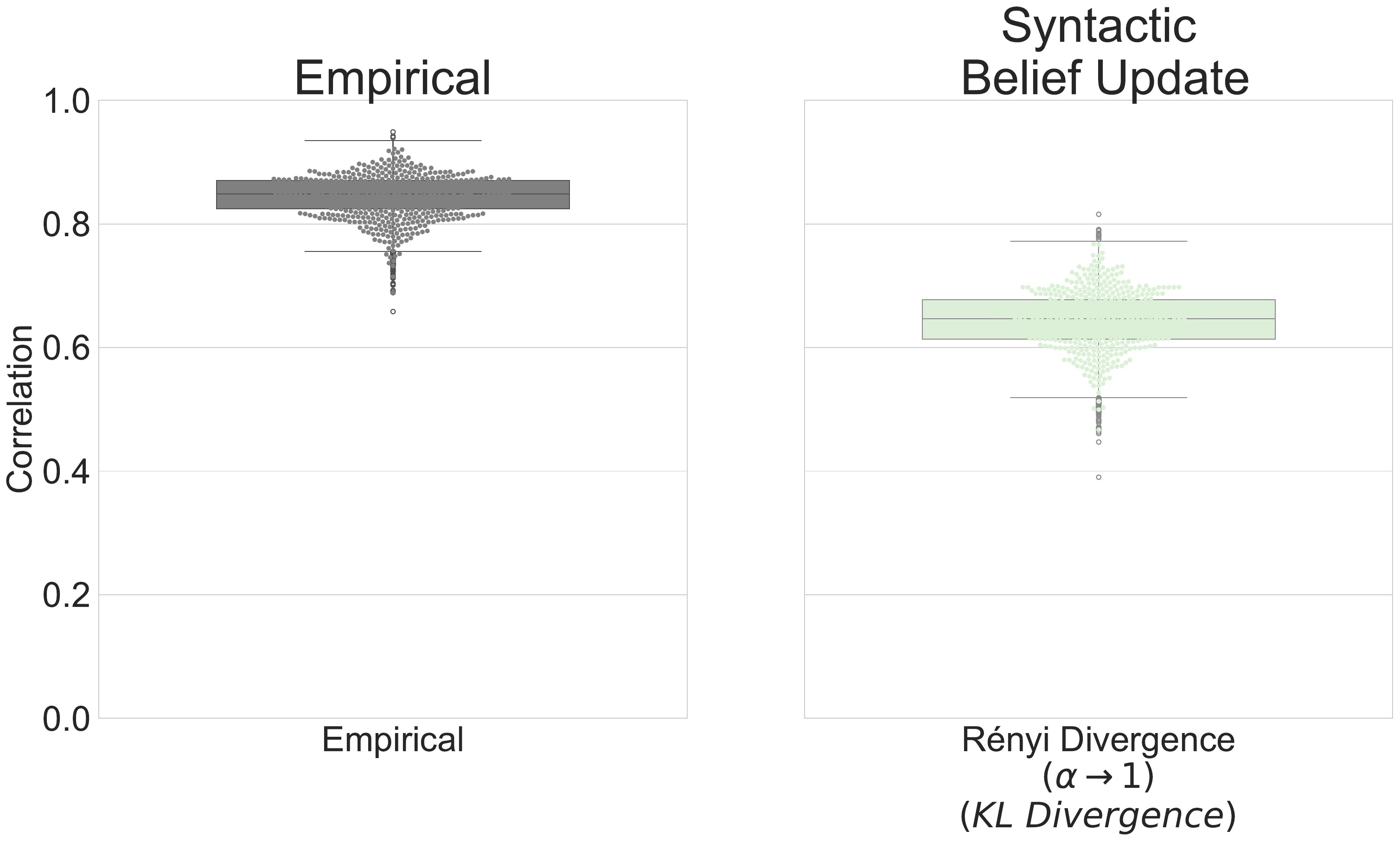}
            \caption{Correlations with Empirical}
        \end{subfigure}
    \end{subfigure}
    \caption{Garden path effects and correlations from predictions derived from Syntactic Belief Update in relation to the true post-critical structure}
    \label{fig:gold}
\end{figure*}
In a similar vein, we preliminarily considered the family of R\'enyi Cross-Entropies as a candidate metric for Syntactic Belief Update and compared them to the R\'enyi Divergence using a linear mixed effects model with the exact same structure as the Bayesian model in Section \ref{sec:app:brms} (Figure \ref{fig:app:alphasweep})

Unlike R\'enyi Divergence, which more closely approximates human reading patterns in both magnitude and correlation as $\alpha$ increases, R\'enyi Cross-Entropy measures perform worse even for $\alpha=2$. If we take the intuition that our Cross-Entropy metrics are a combination of entropy of the post-critical distribution and divergence, then this result suggests that it is ultimately the divergence component that drives garden path processing difficulty. In other words, what matters most for the cost of reanalysis is not the uncertainty of the parser after the critical word, but the distance between its prior belief distribution and the new state it must revise to. Intuitively, this also matches well with theoretical accounts of reanalysis, which emphasize the true structure that a parser must recover to rather than uncertainty after reaching the critical point.

\subsection{KL Divergence against True Post-Critical Structure}
Finally, to control for imperfections in our parser, we also computed Syntactic Belief Updates in relation to the true post-critical structure after seeing the first disambiguating verb. To do this, after sampling a distribution from our incremental parser after it sees the disambiguating verb, we modify the log-potentials of all edges that are fully contained in the current prefix to be highly positive if it's in the non-garden-pathed interpretation, and highly negative if not. To control for any effect on SBU that this sudden narrowing of beliefs has, we do this for both garden path sentences and control sentences. 

Preliminary results (Figure \ref{fig:gold}) using KL Divergence and fit directly on the critical region slowdowns show very close matches in the magnitude of the garden path effect, with our SBU metrics no longer underestimating the true empirical effect for NP/S and MV/RR, and showing similar high correlations with empirical slowdowns across as our high alpha R\'enyi Divergences in mour main results. We tentatively interpret this idealized results as evidence that at least some of the underestimation in our main results are due to model imperfections rather than limitations of our general proposal itself.  

\end{document}